\documentclass[conference,comsoc]{IEEEtran}
\usepackage{cite}

\ifCLASSINFOpdf
  \usepackage[pdftex]{graphicx}
\else
  \usepackage[dvips]{graphicx}
\fi
\usepackage[cmex10]{amsmath}
\usepackage{algorithmic}
\usepackage{array}
\ifCLASSOPTIONcompsoc
  \usepackage[caption=false,font=normalsize,labelfont=sf,textfont=sf]{subfig}
\else
  \usepackage[caption=false,font=footnotesize]{subfig}
\fi
\usepackage{dblfloatfix}
\usepackage{booktabs}
\usepackage{url}
\usepackage{float}
\usepackage{multirow}
\usepackage{enumerate}
\usepackage{amsthm}
\usepackage{threeparttable}
\usepackage{svg}
\usepackage{adjustbox}
\usepackage{booktabs,threeparttable}
\usepackage{array} 
\newcolumntype{C}[1]{>{\centering\arraybackslash}p{#1}} 

\newtheoremstyle{mystyle}
  {}
  {}
  {\itshape}
  {}
  {\bfseries}
  {.}
  { }
  {}
\theoremstyle{mystyle}

\newenvironment{talign*}
 {\let\displaystyle\textstyle\csname align*\endcsname}
 {\endalign}
\usepackage{arydshln}
\usepackage[normalem]{ulem}
\usepackage{hyperref}
\usepackage{amsmath}
\usepackage{xurl}

\begin{document}

%
\title{Explainable AI for Comparative Analysis of Intrusion Detection Models}

\author{
    Pap M. Corea$^{1a}$, Yongxin Liu$^{1a}$, Jian Wang$^{2b}$, Shuteng Niu$^{3c}$, Houbing Song$^{4d}$\\
    $^{1}$Embry-Riddle Aeronautical University, FL 32114 USA,

    $^{2}$University of Tennessee at Martin, TN 38237 USA\\
    $^{3}$Bowling Green State University, OH 43403 USA, $^{4}$University of Maryland, Baltimore County, MD 21250 USA\\

    $^{a}$\{moctarp@my.erau.edu, LIUY11@erau.edu\}, $^{b}$jwang186@utm.edu,
    $^{c}$sniu@bgsu.edu,
    $^{d}$songh@umbc.edu
    
}

\markboth{IEEE Internet of Things Journal,~Vol.~11, No.~4, May~2021}%
{Shell \MakeLowercase{\textit{et al.}}: Bare Demo of IEEEtran.cls for Journals}
\IEEEtitleabstractindextext{%
\begin{abstract}

Explainable Artificial Intelligence (XAI) has become a widely discussed topic, the related technologies facilitate better understanding of conventional black-box models like Random Forest, Neural Networks and etc. However, domain-specific applications of XAI are still insufficient. To fill this gap, this research analyzes various machine learning models to the tasks of binary and multi-class classification for intrusion detection from network traffic on the same dataset using occlusion sensitivity. The models evaluated include Linear Regression, Logistic Regression, Linear Support Vector Machine (SVM), K-Nearest Neighbors (KNN), Random Forest, Decision Trees, and Multi-Layer Perceptrons (MLP). We trained all models to the accuracy of 90\% on the UNSW-NB15 Dataset. We found that most classifiers leverage only less than three critical features to achieve such accuracies, indicating that effective feature engineering could actually be far more important for intrusion detection than applying complicated models. We also discover that Random Forest provides the best performance in terms of accuracy, time efficiency and robustness. Data and code available at \url{https://github.com/pcwhy/XML-IntrusionDetection.git} 
\end{abstract}

}

\IEEEoverridecommandlockouts
\maketitle
\IEEEdisplaynontitleabstractindextext
\IEEEpeerreviewmaketitle

\section{Introduction}

Machine learning (ML) has emerged as a transformative tool in the field of intrusion detection, providing a robust approach to enhancing cybersecurity measures. By leveraging the ability to learn from and adapt to evolving data without explicit programming, ML techniques can effectively identify novel and sophisticated cyber threats. This adaptive capability is crucial in an environment where attackers continuously modify their strategies to evade detection. ML algorithms, including supervised, unsupervised, and reinforcement learning, analyze patterns and anomalies in vast datasets, enabling the prediction and detection of potential intrusions with high accuracy. As such, the application of ML in intrusion detection systems (IDS) represents a significant step forward in developing dynamic, responsive security strategies that can anticipate and mitigate threats in real-time, thus ensuring the integrity and confidentiality of information systems.

Despite the efficacy of machine learning in intrusion detection, the deployment of these technologies raises significant concerns, particularly regarding the opaque nature of certain ML models. Black-box models, such as deep neural networks, often lack transparency in their decision-making processes, making it challenging for cybersecurity professionals to interpret or trust the rationale behind specific detections or classifications \cite{guidotti2018survey}. This uncertainty can complicate compliance with regulatory standards that demand clear audit trails and explainability of security systems. Furthermore, the inability to interpret model decisions can hinder the identification and correction of biases in training data, potentially leading to unfair or ineffective security measures. Such limitations underscore the need for developing more interpretable machine learning models and methods that maintain high detection performance while providing greater transparency and accountability in their operations \cite{arrieta2020explainable}.

In the landscape of explainable AI (XAI), several methods stand out for their ability to render machine learning models more interpretable, especially in critical applications like intrusion detection. LIME (Local Interpretable Model-agnostic Explanations \cite{ribeiro2016should}) is another key technique that approximates the locally predictive behavior of the model around a specific instance, thus providing insights into the decision-making process. SHAP (SHapley Additive exPlanations \cite{lundberg2017unified}) assigns each feature an importance value for a particular prediction, integrating game theory to ensure consistency and accuracy in feature attribution. Grad-CAM (Gradient-weighted Class Activation Mapping \cite{selvaraju2017grad}) uses the gradients of any target concept flowing into the final convolutional layer to produce a coarse localization map highlighting important regions for predictions. Some research even use GradCAM to analyze the potential vulnerabilities within deep neural networks \cite{tan2023noisecam, renkhoff2022exploring}, but such method is only applicable to algbraically differentiable models. Occlusion Sensitivity \cite{zeiler2014visualizing} investigates the influence of different parts of input data on the output by systematically occluding sections of the data and observing the changes in output. This method is particularly useful for identifying which data segments are most critical for decision-making, offering clear visual explanations. Each of these methods offers a different approach to enhance transparency in ML models, but Occlusion Sensitivity is especially valuable for its direct and intuitive visualization capabilities. 

In this paper, we utilize Occlusion Sensitivity to analyze the decision behavior of different machine learning models trained on the UNSW-NB15 Dataset \cite{moustafa2015unsw}, which captures network traffic traces of a hybrid of real modern normal activities and synthetic contemporary attacks. We compare the behavior of various classifiers and ovr findings are as follows:
\begin{itemize}
    \item Most of our models only exploit less than three highly ranked features regardless if feature selection.
    \item The classifiers depends highly on time-dependent features, such as Source or Destination Time to Live (\textit{ttl}), and thus can generate highly domain-specific models. 
    \item Random Forest show the best robust by treating all input features equally, and therefore, it still maintains the best performance while top features are not available.
\end{itemize}

The remainder of this paper is organized as follows: A literature review of related work is presented in Section~\ref{sectRW}. We present the methodology in Section~\ref{sectMM}. Evaluation and discussion are presented in Section~\ref{sectEED} and conclusions in Section~\ref{sectCC}.

\section{Related Work}
\label{sectRW}

Statistical machine learning is characterized by its ability to model complex data through probabilistic approaches, enabling systems to make predictions or decisions based on data analysis, helping in tasks like classification, regression, and clustering under uncertainty. Statistical machine learning has played a pivotal role in advancing network intrusion detection systems (NIDS), offering diverse approaches for identifying and mitigating cyber threats. For instance, Barbara et al. \cite{barbara2001adam} utilized data mining algorithms to develop the ADAM project, a real-time anomaly detection system. Another work, done by Tavallaee et al. \cite{tavallaee2009detailed}, presented an improved KDD dataset for benchmarking intrusion detection algorithms. Similarly, Kruegel and Vigna \cite{kruegel2003anomaly} explored anomaly detection using sequences of system calls, enhancing the detection accuracy of host-based IDS. Additionally, Thaseen and Kumar \cite{thaseen2017intrusion} integrated SVM classifiers with feature reduction techniques to efficiently handle high-dimensional data in network traffic. Despite its wide adaption, Statistical learning models needs intensive human efforts in feature engineering and can be susceptible to overfitting, particularly when the data has high variance or the model is too complex, leading to poor generalization on new, unseen data.

Compared with Statistical Machine Learning, Deep learning has increasingly become a pivotal approach, providing robust mechanisms to detect sophisticated cyber threats. Deep learning models, primarily due to their ability to learn complex patterns without hard effort in feature engineering from large volumes of data, have shown significant promise in distinguishing between normal traffic and potential threats with high accuracy. Yin et al. \cite{yin2017deep}, demonstrated their effectiveness in capturing spatial features within network traffic. Similarly, Kim et al. \cite{kim2016long} employed Recurrent Neural Networks (RNNs), particularly Long Short-Term Memory (LSTM) networks, to analyze temporal features of traffic data for anomaly detection. Further, Javaid et al. \cite{javaid2016deep} explored the use of Self-Taught Learning (STL), a hybrid model that combines deep learning with sparse coding to enhance feature learning in an unsupervised manner. While deep learning offers substantial improvements in network intrusion detection, it also presents several challenges. One major drawback is the requirement for vast amounts of labeled training data, which is expensive and time-consuming to gather in the cybersecurity domain. Additionally, deep learning models are often seen as "black boxes," providing limited interpretability regarding their decision-making processes, which can be a critical shortfall in security applications where understanding the rationale behind decisions is essential.
\begin{figure}[b]
    \centering
    \includegraphics[width=0.6\linewidth]{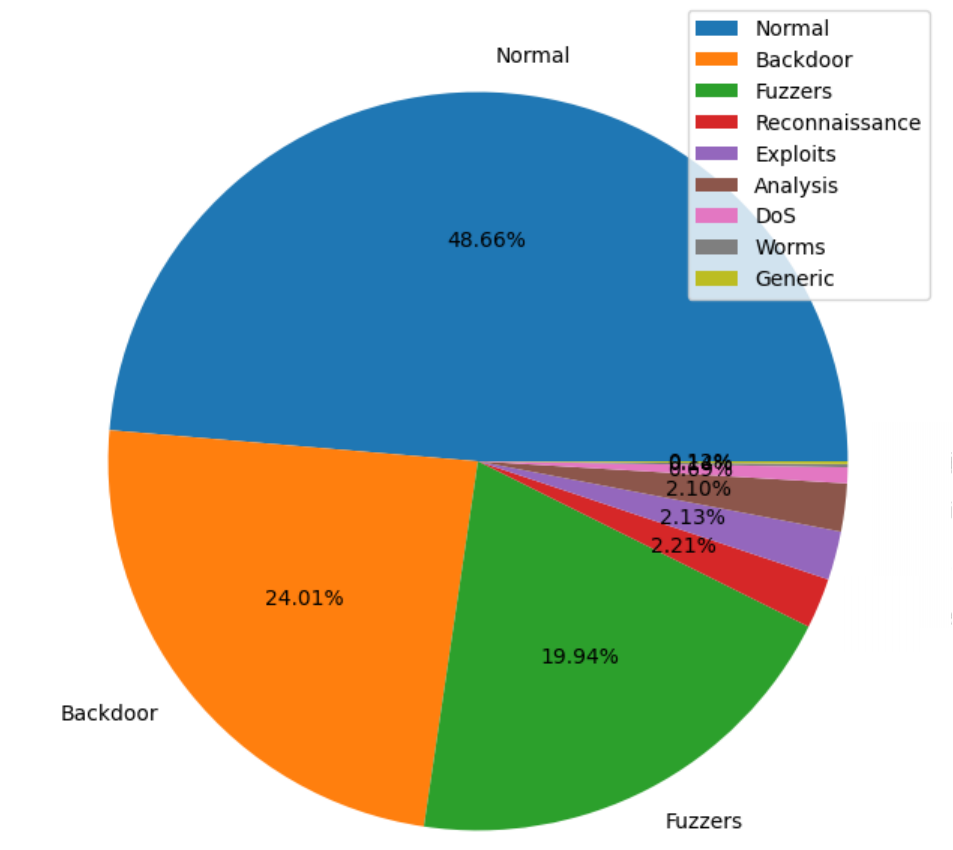}
    \caption{Distribution of intrusion attack categories after data preprocessing.}
    \label{FigAttackTypeDistribution}
\end{figure}
  
\begin{figure}
    \centering
    \includegraphics[width=\linewidth]{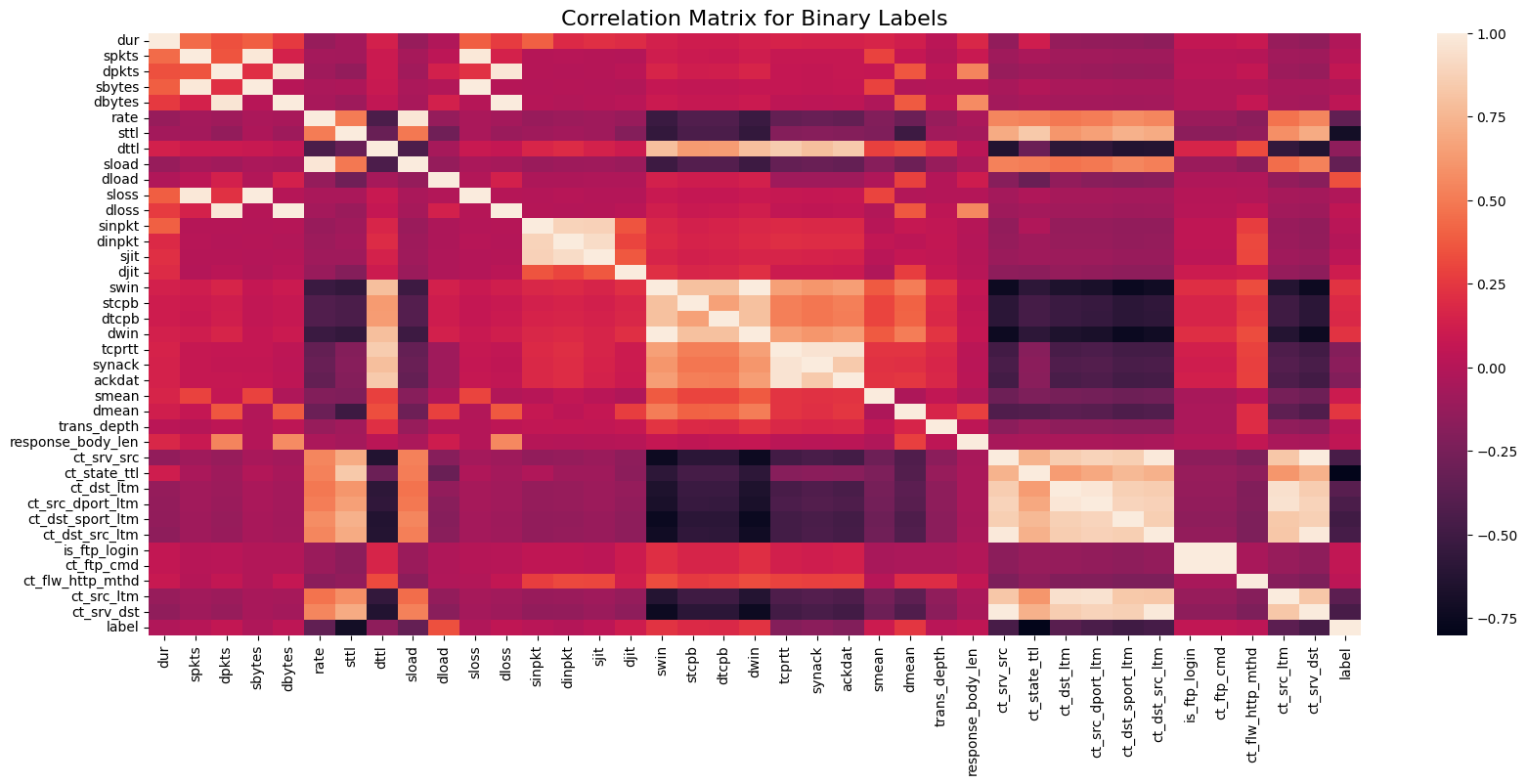}
    \vspace{-8mm}
    \caption{Feature correlation matrix}
    \label{figFeatureCorr}
\end{figure}

\begin{figure}
    \centering
    \includegraphics[width=0.7\linewidth]{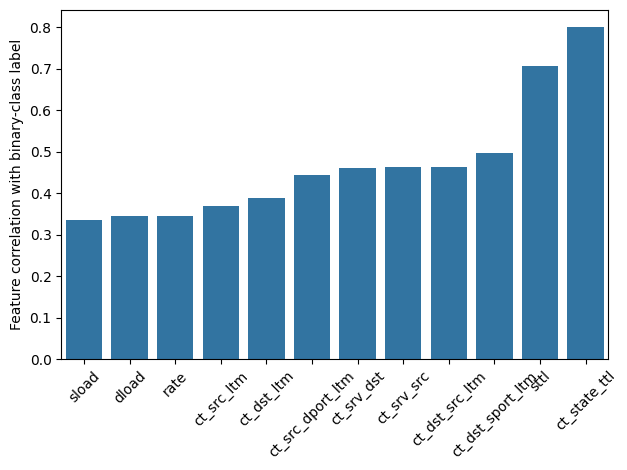}
    \caption{Selected features for binary classifiers.}
    \label{figFeatureCorrBinLabel}
\end{figure}
\begin{figure}
    \centering
    \includegraphics[width=0.65\linewidth]{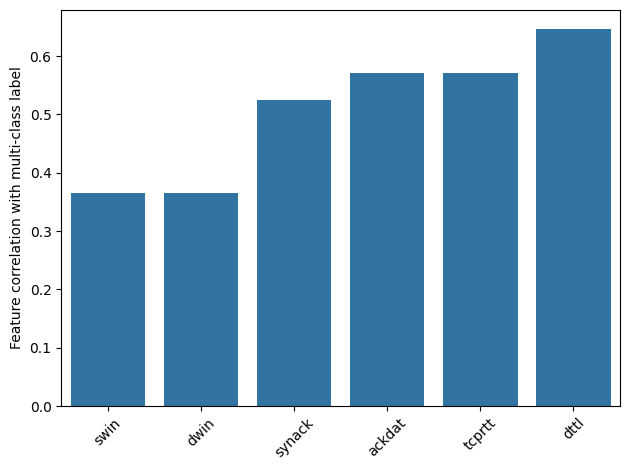}
    \caption{Selected features for multi-class classifiers.}
    \label{figFeatureCorrMulLabel}
\end{figure}  
\begin{figure*}[b]
    \centering
    \includegraphics[width=\linewidth]{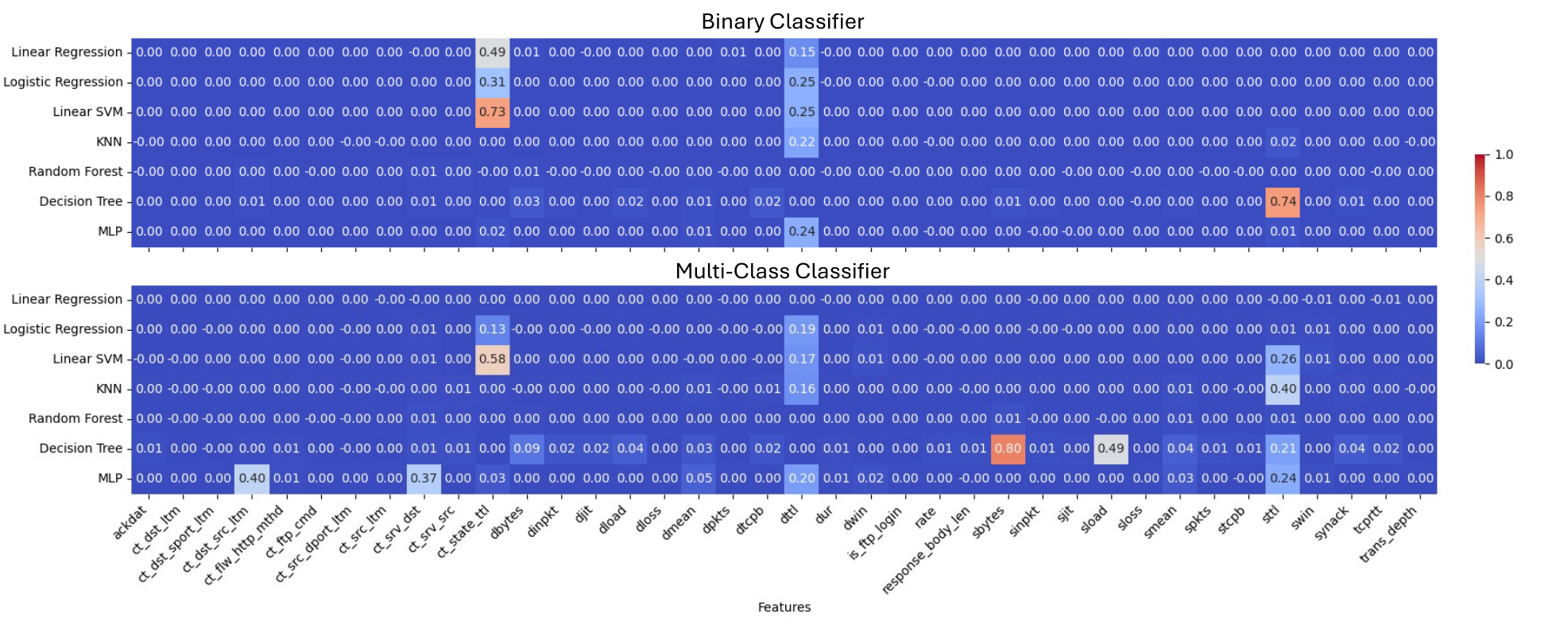}
    \caption{Feature sensitivity of intrusion detection model classifiers trained with complete features.}
    \label{figFullFeatureSensitivity}
\end{figure*}
The growing interest in making machine learning models, especially those applied to network intrusion detection, more interpretable and trustworthy, has spurred the development of explainable artificial intelligence (XAI) approaches in this domain. For instance, Sauka et al. \cite{sauka2022adversarial} developed an adversarial robust and explainable intrusion detection system using deep learning, emphasizing the enhancement of model transparency and robustness. Patil and colleagues \cite{patil2022explainable} proposed a machine learning-based intrusion detection system that highlights the potential of explainability in security applications, focusing on demystifying the black-box nature of complex models. Keshk et al. \cite{keshk2023explainable} introduced an explainable deep learning framework specifically tailored for IoT networks, underscoring the critical need for clarity in automated security systems within such environments. Furthermore, Wang et al. \cite{wang2020explainable} and Barnard et al. \cite{barnard2022robust} have contributed significantly by integrating techniques like SHAP (SHapley Additive exPlanations) to elucidate the decision-making processes of their intrusion detection models, thus facilitating a better understanding and trust among network security personnel.

Explainable machine learning (XAI) models for network intrusion detection often face challenges such as increased computational complexity and potentially reduced performance due to the overhead of generating explanations. Additionally, while providing transparency, the explanations themselves may be too technical or abstract for non-specialist users, limiting their practical usefulness in real-world security applications where clear and actionable insights are required. Such limitations motivate us to use Explainable AI method to perform a comparative analysis on the behaviors of different machine learning-enabled intrusion detectors on the same dataset.

\section{Methodology}
\label{sectMM}
We use Occlusion Sensitivity to analyze the behavior of different machine learning models on UNSW-NB15 dataset. We want to see if the trained machine learning model we use for IDS could unintentionally become biased towards specific features.
\subsection{Data Preprocessing}
The UNSW-NB15 Dataset \cite{moustafa2015unsw} contains 175,341 entries across 45 distinct columns. We conduct the following data preprocessing steps:

\begin{itemize}
    \item \textit{Removal of incomplete records: }we remove records containing missing values resulting in a reduced dataset of 81,173 entries. An overview of intrusion category is given in Figure~\ref{FigAttackTypeDistribution}, the prevalence of ’Normal’ traffic at 48.66\%, followed by significant portions of ’Generic’ at 24.01\%, ’Fuzzers’ at 19.94\%, and smaller fractions for ’Backdoor’, ’Analysis’, ’Exploits’, ’Reconnaissance’, ’DoS’, and ’Worms’.
    \item \textit{Encoding categorical features: }we convert categorical features into one-hot encoding.
    \item \textit{Scaling and Normalization: }We re-scale numerical values to the range of $[0, 1]$.
    \item \textit{Feature Selection: }we remove features that have less than 0.3 of correlation with the classification label. The correlation matrix of features is given in Figure~\ref{figFeatureCorr}. The selected features for both binary and multi-class classifiers as well as their Correlation Coefficients are in Figures \ref{figFeatureCorrBinLabel} and \ref{figFeatureCorrMulLabel}. To evaluate the impact of such feature selection criteria, we also train and analyze the models without feature selection as for comparison.
    \item \textit{Data Synthetic and Model Training: }We divided the dataset randomly to compose training (80\%) and test (20\%) set, our stopping criteria for model training are either reach 90\% of classification accuracy or improvement less than 1\% after the latest epoch.

\end{itemize}

\subsection{Binary Classifier Analysis}
The binary classification task distinguishes between normal network behavior (non-intrusive) and abnormal behavior (intrusive). The models employed for this task are imported directly from scikit-learn library with default configurations, they are: Linear Regression, Logistic Regression, Linear SVM, K-Nearest Neighbor, Random Forest, Decision Tree, and MLP. 

\subsection{Multi-Class Classification Analysis}
The multi-class intrusion detection models utilize the same suite of models to predict various attack categories such as DoS, Exploits, Fuzzers, and others. Similar performance metrics have been calculated for the multi-class models to evaluate their effectiveness in distinguishing between the different attack categories. Additionally, occlusion sensitivity has been implemented to identify the most influential features for the predictions. 

\section{Evaluation \& Discussion}
\label{sectEED}
\subsection{Binary Classifiers}
\begin{figure}[]
    \centering
    \includegraphics[width=\linewidth]{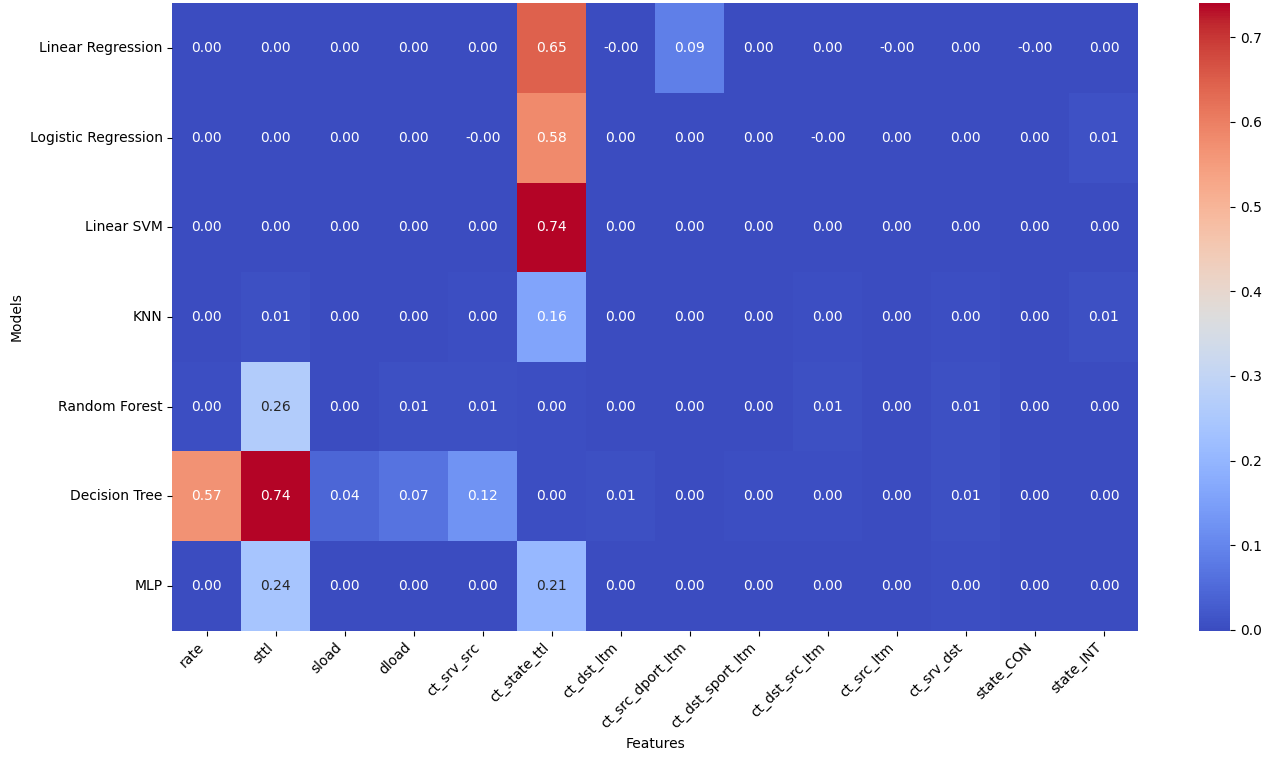}
    \caption{Feature sensitivity of binary intrusion detection model classifiers trained with selected features.}
    \label{figFeatureImportanceBinClassifier}
\end{figure}

The feature sensitivity with respect to classification accuracy degradation of binary classifiers are given in Figures \ref{figFullFeatureSensitivity} and ~\ref{figFeatureImportanceBinClassifier}. As depicted, most binary models are extremely sensitive to less than three top features, in particular, Decision Tree model is extremely sensitive to even single feature occlusion. Meanwhile, Multi-Layer Feed-Forward Neural Network and Random Forest models exploit more features than other models. We adjusted the L2 regularization coefficient of MLP model from 0.0001 to 0 and we did not observe significant differences. A possible explanation is that the neural network only leverages a few highly important features and thus develops a sparse internal structure which is not sensitive to the L2 regularization.

\begin{figure}[h]
    \centering
    \includegraphics[width=\linewidth]{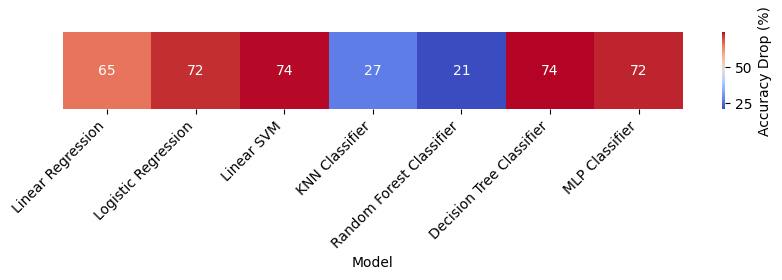}
    \caption{Accuracy degradation after masking the Top-2 features.}
    \label{figDualFeatureMasking}
\end{figure}

We masked the Top-2 features of the binary classifiers to analyze the performance degradation of classifiers, depicted in Figure~\ref{figDualFeatureMasking}. We found that only Random Forest and K-NN classifiers maintain the most insignificant performance degradation.

\subsection{Multi-Class Classifiers}
The feature sensitivity results of multi-class classifiers also indicates that models utilize more features than in binary classifier when there's no feature selection procedure. Conversely, when feature selection is performed, most classifiers indicates the \textit{ttl}-related features are more critical for classification accuracy. Interestingly, Random Forest is not sensitive to single feature masking in both binary and multi-class scenarios, similar to binary classification scenario, Decision Tree model is extremely sensitive to even single feature occlusion.
\begin{figure}[h]
    \centering
    \includegraphics[width=\linewidth]{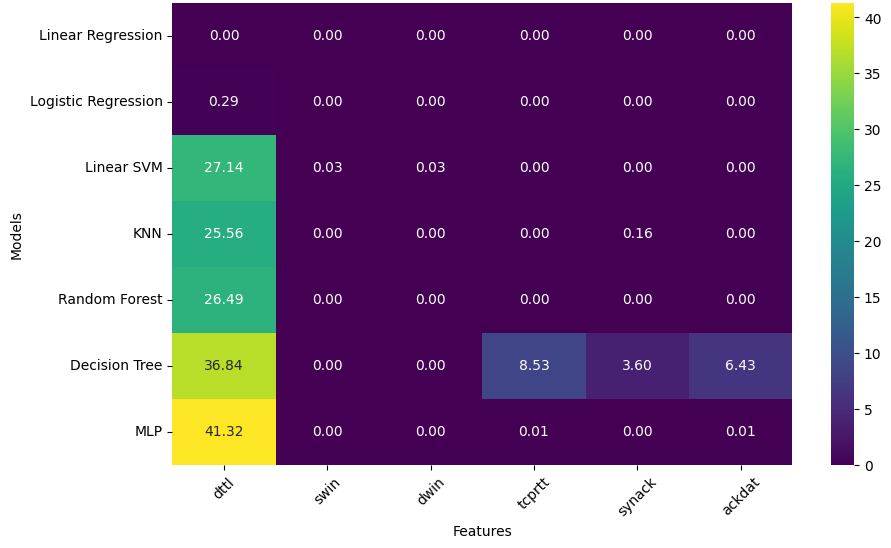}
    \caption{Feature sensitivity w.r.t. classification accuracy of multi-class intrusion detection model.}
    \label{figFIMultiClassifiers}
\end{figure}

\begin{figure}[h]
    \centering
    \includegraphics[width=\linewidth]{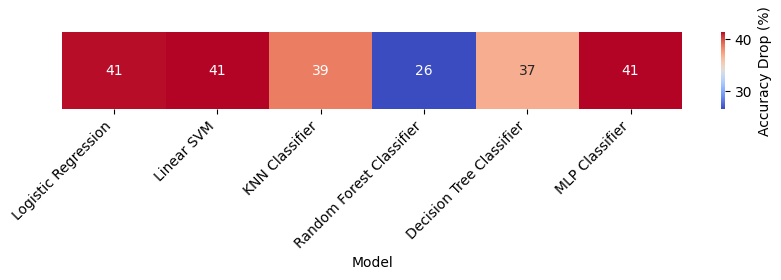}
    \caption{Accuracy degradation after masking the Top-2 features}
    \label{figDualFeatureMaskingMC}
\end{figure}

Interestingly, if we mask out the top-2 most important features, as depicted in Figure~\ref{figDualFeatureMaskingMC}, the models performance degradation may not be as significant as in binary classifiers as in Figure~\ref{figDualFeatureMasking}. We  still find that Random Forest still has the best robustness when the top-2 features are mask-out. Moreover, we found that the classifiers rely highly on the time-dependent features, such as \textit{sttl} and \textit{dttl}, simply means that the all these models may face challenges or become useless if they are ported to different application scenarios.
\begin{figure*}
    \centering
    \includegraphics[width=\linewidth]{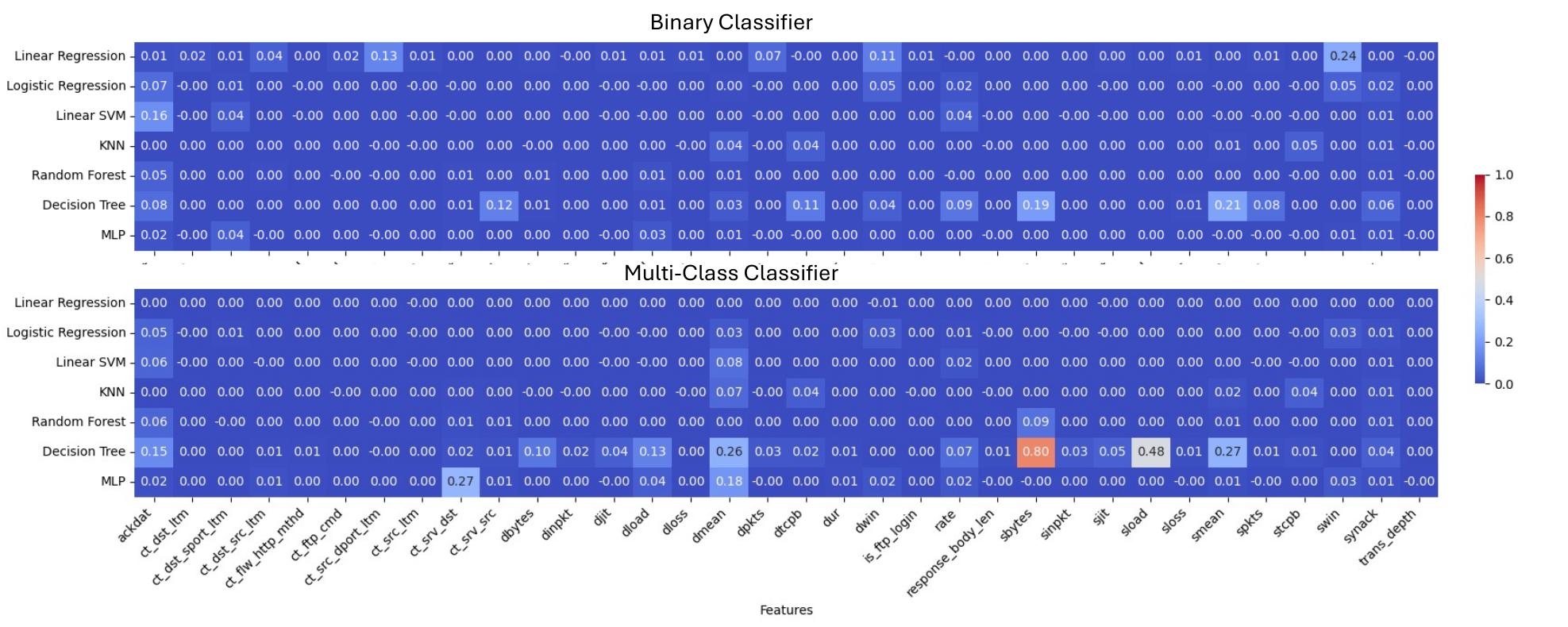}
    \caption{Feature sensitivity of intrusion detection model classifiers trained without the top features.}
    \label{figFullFeatureSensitivityWithoutTTL}
\end{figure*}

To compare the feature sensitivity of the models, we re-train all the models with the top-3 \textit{TTL}-related features removed and derive the feature sensitivities in Figure~\ref{figFullFeatureSensitivityWithoutTTL}. Compare with Figure~\ref{figFullFeatureSensitivity}, the models utilize more features while the Decision Tree model still has a strong bias towards specific features.

\subsection{Model Overhead Comparison}

\begin{figure}[h]
    \centering
    \includegraphics[width=0.9\linewidth]{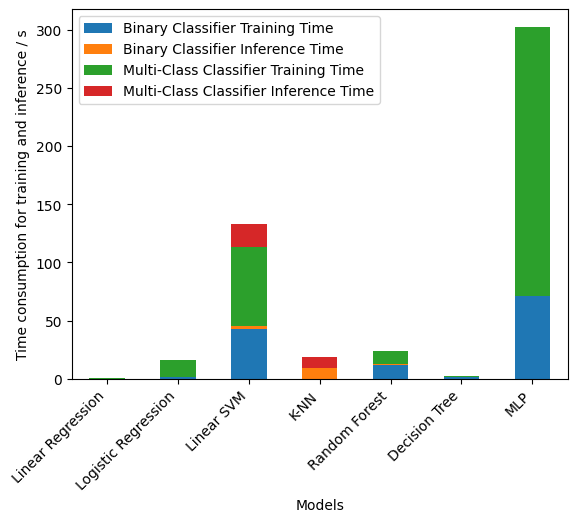}
    \caption{Comparison of model overhead.}
    \label{figModelOverhead}
\end{figure}

We compare the time consumption of deriving all the models considering full feature set as in Figure~\ref{figModelOverhead}. Our experiment is done in standard Google Colab environment with Intel(R) Xeon(R) CPU at 2.20GHz and 12.7GB of RAM, as depicted, Random Forest becomes the best model for the UNSW-NB15 dataset by fully utilizing the features and providing the best efficiency.

\section{Conclusion}
This paper utilizes Occlusion Sensitivity method for a comparative study on feature importance of various machine learning models for network intrusion detection on UNSW-NB15 dataset. We found that most machine learning models, including Neural Network model exploit few critical features to make decisions and users have to mask out critical features to let model focus on other useful features. In the meantime, Random Forest is the only model that treat all input features equally. Our further experiment also reveal that Random Forest is more veratile than neural network models such as MLP by proving similar performance with better robustness and significantly less training time. Our finding also indicates that explainable AI-guided feature engineering could be a promising approach for deriving robust model while maintain uncompromising performances.



Our future direction includes improving airspace ATC workload assessment by considering metrics beyond delayed and total flights. Additionally, we plan to explore neural networks' potential in generating comprehensive airspace configuration plans.

\label{sectCC}

\section*{Acknowledgment}

This research was supported by the Center for Advanced Transportation Mobility (CATM), USDOT Grant No. 69A3551747125, 270128BB(AWD00237), the U.S. National Science Foundation under Grant No.2231629, 2142514 and Grant No.2309760 and the USDOT Tier-1 University Transportation Center (UTC) Transportation Cybersecurity Center for Advanced Research and Education (CYBER-CARE) (Grant No. 69A3552348332).

\bibliographystyle{IEEEtran}
\bibliography{IPCCC2022.bib}

\begin{thebibliography}{10}
\providecommand{\url}[1]{#1}
\csname url@samestyle\endcsname
\providecommand{\newblock}{\relax}
\providecommand{\bibinfo}[2]{#2}
\providecommand{\BIBentrySTDinterwordspacing}{\spaceskip=0pt\relax}
\providecommand{\BIBentryALTinterwordstretchfactor}{4}
\providecommand{\BIBentryALTinterwordspacing}{\spaceskip=\fontdimen2\font plus
\BIBentryALTinterwordstretchfactor\fontdimen3\font minus
  \fontdimen4\font\relax}
\providecommand{\BIBforeignlanguage}[2]{{%
\expandafter\ifx\csname l@#1\endcsname\relax
\typeout{** WARNING: IEEEtran.bst: No hyphenation pattern has been}%
\typeout{** loaded for the language `#1'. Using the pattern for}%
\typeout{** the default language instead.}%
\else
\language=\csname l@#1\endcsname
\fi
#2}}
\providecommand{\BIBdecl}{\relax}
\BIBdecl

\bibitem{guidotti2018survey}
R.~Guidotti, A.~Monreale, S.~Ruggieri, F.~Turini, F.~Giannotti, and
  D.~Pedreschi, ``A survey of methods for explaining black box models,''
  \emph{ACM computing surveys (CSUR)}, vol.~51, no.~5, pp. 1--42, 2018.

\bibitem{arrieta2020explainable}
A.~B. Arrieta, N.~D{\'\i}az-Rodr{\'\i}guez, J.~Del~Ser, A.~Bennetot, S.~Tabik,
  A.~Barbado, S.~Garc{\'\i}a, S.~Gil-L{\'o}pez, D.~Molina, R.~Benjamins
  \emph{et~al.}, ``Explainable artificial intelligence (xai): Concepts,
  taxonomies, opportunities and challenges toward responsible ai,''
  \emph{Information fusion}, vol.~58, pp. 82--115, 2020.

\bibitem{ribeiro2016should}
M.~T. Ribeiro, S.~Singh, and C.~Guestrin, ``" why should i trust you?"
  explaining the predictions of any classifier,'' in \emph{Proceedings of the
  22nd ACM SIGKDD international conference on knowledge discovery and data
  mining}, 2016, pp. 1135--1144.

\bibitem{lundberg2017unified}
S.~M. Lundberg and S.-I. Lee, ``A unified approach to interpreting model
  predictions,'' \emph{Advances in neural information processing systems},
  vol.~30, 2017.

\bibitem{selvaraju2017grad}
R.~R. Selvaraju, M.~Cogswell, A.~Das, R.~Vedantam, D.~Parikh, and D.~Batra,
  ``Grad-cam: Visual explanations from deep networks via gradient-based
  localization,'' in \emph{Proceedings of the IEEE international conference on
  computer vision}, 2017, pp. 618--626.

\bibitem{tan2023noisecam}
W.~Tan, J.~Renkhoff, A.~Velasquez, Z.~Wang, L.~Li, J.~Wang, S.~Niu, F.~Yang,
  Y.~Liu, and H.~Song, ``Noisecam: Explainable ai for the boundary between
  noise and adversarial attacks,'' \emph{arXiv preprint arXiv:2303.06151},
  2023.

\bibitem{renkhoff2022exploring}
J.~Renkhoff, W.~Tan, A.~Velasquez, W.~Y. Wang, Y.~Liu, J.~Wang, S.~Niu, L.~B.
  Fazlic, G.~Dartmann, and H.~Song, ``Exploring adversarial attacks on neural
  networks: An explainable approach,'' in \emph{2022 IEEE International
  Performance, Computing, and Communications Conference (IPCCC)}.\hskip 1em
  plus 0.5em minus 0.4em\relax IEEE, 2022, pp. 41--42.

\bibitem{zeiler2014visualizing}
M.~D. Zeiler and R.~Fergus, ``Visualizing and understanding convolutional
  networks,'' in \emph{Computer Vision--ECCV 2014: 13th European Conference,
  Zurich, Switzerland, September 6-12, 2014, Proceedings, Part I 13}.\hskip 1em
  plus 0.5em minus 0.4em\relax Springer, 2014, pp. 818--833.

\bibitem{moustafa2015unsw}
N.~Moustafa and J.~Slay, ``Unsw-nb15: a comprehensive data set for network
  intrusion detection systems (unsw-nb15 network data set),'' in \emph{2015
  military communications and information systems conference (MilCIS)}.\hskip
  1em plus 0.5em minus 0.4em\relax IEEE, 2015, pp. 1--6.

\bibitem{barbara2001adam}
D.~Barbar{\'a}, J.~Couto, S.~Jajodia, and N.~Wu, ``Adam: a testbed for
  exploring the use of data mining in intrusion detection,'' \emph{ACM Sigmod
  Record}, vol.~30, no.~4, pp. 15--24, 2001.

\bibitem{tavallaee2009detailed}
M.~Tavallaee, E.~Bagheri, W.~Lu, and A.~A. Ghorbani, ``A detailed analysis of
  the kdd cup 99 data set,'' in \emph{2009 IEEE symposium on computational
  intelligence for security and defense applications}.\hskip 1em plus 0.5em
  minus 0.4em\relax Ieee, 2009, pp. 1--6.

\bibitem{kruegel2003anomaly}
C.~Kruegel and G.~Vigna, ``Anomaly detection of web-based attacks,'' in
  \emph{Proceedings of the 10th ACM conference on Computer and communications
  security}, 2003, pp. 251--261.

\bibitem{thaseen2017intrusion}
I.~S. Thaseen and C.~A. Kumar, ``Intrusion detection model using fusion of
  chi-square feature selection and multi class svm,'' \emph{Journal of King
  Saud University-Computer and Information Sciences}, vol.~29, no.~4, pp.
  462--472, 2017.

\bibitem{yin2017deep}
C.~Yin, Y.~Zhu, J.~Fei, and X.~He, ``A deep learning approach for intrusion
  detection using recurrent neural networks,'' \emph{Ieee Access}, vol.~5, pp.
  21\,954--21\,961, 2017.

\bibitem{kim2016long}
J.~Kim, J.~Kim, H.~L.~T. Thu, and H.~Kim, ``Long short term memory recurrent
  neural network classifier for intrusion detection,'' in \emph{2016
  international conference on platform technology and service (PlatCon)}.\hskip
  1em plus 0.5em minus 0.4em\relax IEEE, 2016, pp. 1--5.

\bibitem{javaid2016deep}
A.~Javaid, Q.~Niyaz, W.~Sun, and M.~Alam, ``A deep learning approach for
  network intrusion detection system,'' in \emph{Proceedings of the 9th EAI
  International Conference on Bio-inspired Information and Communications
  Technologies (formerly BIONETICS)}, 2016, pp. 21--26.

\bibitem{sauka2022adversarial}
K.~Sauka, G.-Y. Shin, D.-W. Kim, and M.-M. Han, ``Adversarial robust and
  explainable network intrusion detection systems based on deep learning,''
  \emph{Applied Sciences}, vol.~12, no.~13, p. 6451, 2022.

\bibitem{patil2022explainable}
S.~Patil, V.~Varadarajan, S.~M. Mazhar, A.~Sahibzada, N.~Ahmed, O.~Sinha,
  S.~Kumar, K.~Shaw, and K.~Kotecha, ``Explainable artificial intelligence for
  intrusion detection system,'' \emph{Electronics}, vol.~11, no.~19, p. 3079,
  2022.

\bibitem{keshk2023explainable}
M.~Keshk, N.~Koroniotis, N.~Pham, N.~Moustafa, B.~Turnbull, and A.~Y. Zomaya,
  ``An explainable deep learning-enabled intrusion detection framework in iot
  networks,'' \emph{Information Sciences}, vol. 639, p. 119000, 2023.

\bibitem{wang2020explainable}
M.~Wang, K.~Zheng, Y.~Yang, and X.~Wang, ``An explainable machine learning
  framework for intrusion detection systems,'' \emph{IEEE Access}, vol.~8, pp.
  73\,127--73\,141, 2020.

\bibitem{barnard2022robust}
P.~Barnard, N.~Marchetti, and L.~A. DaSilva, ``Robust network intrusion
  detection through explainable artificial intelligence (xai),'' \emph{IEEE
  Networking Letters}, vol.~4, no.~3, pp. 167--171, 2022.

\end{thebibliography}

\end{document}